\documentclass[12pt]{article}
\usepackage{amssymb,amsmath,cite,bm}
\usepackage[dvips]{graphicx,color}
\usepackage{psfrag,subfigure}
\newtheorem{proposition}{Proposition}

\newtheorem{assumption}{Assumption}

\begin{document}

\title{\bf Six-DOF Spacecraft Dynamics Simulator For Testing Translation and Attitude Control}

\author{Farhad Aghili\thanks{email: faghili@encs.concordia.ca}}

\date{}

\maketitle

\begin{abstract}
This paper presents a method to control a manipulator system
grasping a rigid-body payload so that the motion of the combined
system in consequence of external applied forces to be the same as
another free-floating rigid-body (with  different inertial
properties). This allows zero-g emulation of a scaled spacecraft
prototype  under the test in a 1-g laboratory environment. The
controller consisting of motion feedback and force/moment feedback
adjusts the motion of the test spacecraft  so as to match that of
the flight spacecraft, even if the latter has flexible appendages
(such as solar panels) and the former is rigid. The stability of the
overall system  is analytically investigated, and the results show
that the system remains stable provided that the inertial properties
of two spacecraft are different and that an upperbound on the norm
of the inertia ratio of the payload to manipulator is respected.
Important practical issues such as calibration and sensitivity
analysis to sensor noise and quantization are also presented.
\end{abstract}


\section{Introduction and Motivation}
\label{sec_introduction}

Ground testbed facilities have been used for spacecraft control
hardware/software verification since various space programs began
half a century ago \cite{Aghili-Namvar-2008,Aghili-Namvar-Vukovich-2006,Schwartz-Peck-Hall-2003,Piedboeuf-deCarufel-Aghili-Dupuis-1999}. Due to the high
cost of launch and operations associated with on-orbit repair, a
spacecraft must operate reliably  once it is placed in orbit.
Therefore, realistic testing of  spacecraft prior to launch, ideally
with all hardware/software in place, ought to be undertaken to
ensure that the spacecraft functions as intended. One of the
challenges of this approach is that testing must take place in a 1-g
environment, whereas the actual system will eventually operate in a
zero-g environment. This has motivated building of testbed
facilities in various government and university laboratories for the
analysis and testing of spacecraft.

Gas-jet thrusters and reaction/momentum wheels are commonly employed
as actuators for spacecraft attitude and/or translation control
\cite{Graiffin-French-1991}. Simulation is widely used for
characterizing the functional behavior of spacecraft control systems
\cite{Wiesel-1989,Thomson-1986}. This approach may be inadequate and
it should be highly desirable to be able to test and validate system
performance based on the behavior of actual sensors and actuators,
which are difficult to model \cite{Wertz-1978,Aghili-2010n,Aghili-Parsa-2008b,Graiffin-French-1991}.
There are many technologies to address the problem of reproducing
the micro-gravity space environment, such as air bearings,
underwater test tanks, free-fall tests, and magnetic suspension
systems. However, of these, only air bearings have proven useful for
testing spacecraft. Achieving weightlessnees by using natural
buoyancy facilities, i.e., water tank, has been used extensively for
astronaut training. However,  a functional spacecraft can not be
submerged in the water, and in addition viscous damping does not
allow a perfect force-free environment. A free-fall test through
flying parabolas in aircraft can achieve zero-g in a 3-D
environment. But only for brief periods. Magnetic suspension systems
provide only a low force-torque dynamic environment with a small
range of motion. Air-bearing tables (also known as planar
air-bearings) \cite{Yoshida-1995,Aghili-2005b} and spherical
air-bearings \cite{Schwartz-Peck-Hall-2003} are commonly used for
ground-based testbeds for testing the translation and attitude
control systems of a spacecraft.

An emulation of zero-g
translational motion can be achieved by an air-bearing table on
which a spacecraft translates on a surface perpendicular to the
gravity vector while being floated on a cushion of compressed air
with almost no resistance. This technique has been used for testing
various space systems such as formation flying
\cite{Corazzini-Robertson-1996,Choset-Kortenkamp-1999}, free-flying
space robots \cite{Schubert-How-1997,Aghili-2010p}, orbital
rendezvous and docking \cite{Matunaga-Yoshihara-2000,Aghili-2012b}, capturing
mechanisms of spacecraft \cite{Kawamoto-Matsumoto-Wakabayashi-2001,Aghili-2011k},
and free-flying inspection vehicles \cite{Choset-Kortenkamp-1999,Aghili-Parsa-2007b}, and space robotics \cite{Aghili-Dupuis-Piedboeuf-deCarufel-1999,Aghili-p1,Piedboeuf-deCarufel-Aghili-Dupuis-1999,Doyon-Piedboeuf-Aghili-Gonthier-Martin-2003,Aghili-Parsa-2007b,Piedboeuf-Aghili-Doyon-Martin-2002,Aghili-Piedboeuf-2002,Aghili-Parsa-2009b,Aghili-Su-2012}.
Although the air-bearing table system can be utilized to test some
physical components of spacecraft control systems including the
sensors and actuators, this system is limited to a two-dimensional
planar environment. Spherical air-bearings have been used for
spacecraft attitude determination and control hardware/software
verification for many years \cite{Schwartz-Peck-Hall-2003}. The
earliest development and design of a satellite simulator based on
spherical air-bearing with three axes of rotation was reported in
\cite{Bachofer-Seaman-1964}, and has now evolved into  modern
testbed facilities
\cite{Colebank-Jones-Nagy-Pollak-1999,Miller-Saenz-Wertz-2000,Peck-Miller-2003}.
A spherical air-bearing yields minimum friction and hence offers a
nearly torque-free environment if the center of mass is coincident
with the bearing's center of rotation. The main problem with the air
bearing system is the limited range of motion resulting from
equipment being affixed to the bearing \cite{Peck-Miller-2003}.
Also, spherical air-bearings are not useful for simulating
spacecraft having flexible appendages, because the location of the
center-of-mass of such spacecraft is not fixed. Although one can
envisage combining the two air-bearing technologies in a testbed for
reproducing both the rotational and translational motions, complete
freedom in all six rigid degrees-of-freedom is still technically
difficult to achieve \cite{Schwartz-Peck-Hall-2003}.

Motion table testing systems allow the incorporation of real sensors
of a satellite such as gyros and star trackers in
Hardware-In-The-Loop (HIL) simulation loops.
However, actuators such as reaction wheels or gas-jet thrusters have
been simulated. The main idea in  HIL simulation is that of
incorporating a part of real hardware in the simulation loop during
the system development \cite{Bacic-2005}. Rather than testing the
control algorithm on a purely mathematical model of the system, one
can use real hardware in the simulation loop
\cite{Bacic-2005,Aghili-Namvar-Vukovich-2006}. This allows for
detailed measurement for accurate performance assessment of the
system under the test. The concept of the HIL methodology has also
been utilized for design and implementation of various laboratory
testbeds to study the dynamic coupling between a space-manipulator
and its host spacecraft operating in free space
\cite{Sato-Ejiri-Iada-Fujii-1991,Aghili-2016c,Dubowsky-Durfee-Kuklinski-1994,Aghili-2010f,Yoshida-1995,Akima-Tarao-Uchiyama-1999,Aghili-2006,Tarao-Inohira-Uchiyama-2000,Aghili-2009b,yoshida-nakanishi-ueno-2004,Aghili-2006b}.
A system called the Vehicle Emulation System Model II (VES II)
permits the experimental evaluation of planning and control
algorithm for mobile terrestrial and space robot systems by using
the so-called "admittance control"
\cite{Dubowsky-Durfee-Kuklinski-1994}. Similar concepts have been
also pursued by other space agencies such as DLR
\cite{Krenn-Schafer-1999}, NASA and CSA 
\cite{Ananthakrishnan-Teders-Alder-1996,Aghili-p2,Aghili-Parsa-2008b,Aghili-2019c} for different applications.

The existing impedance-controller based HIL simulators only
compensate for the effect of gravity wrench on the force/moment
measurement, while the effect of the payload's inertial forces (the
test spacecraft in our case) has not been taken into account. Heavy
payloads, however, not only changes the manipulator dynamics but
also, incorporate significant inertial as well as gravitational
force components into the measurement that can fail a conventional
impedance controller to achieve the desired dynamics.

\begin{figure}[t]
\centering\includegraphics[width=8cm]{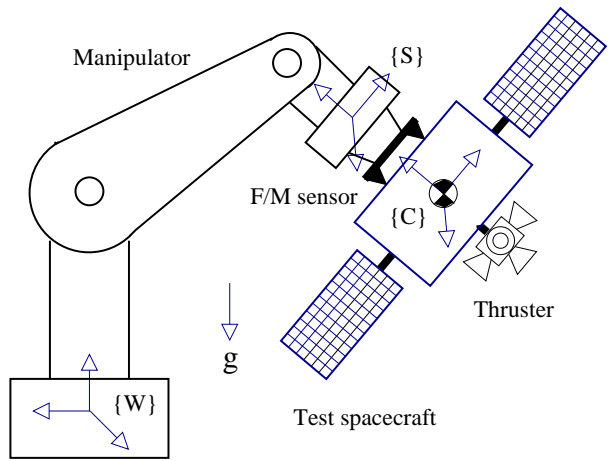}
\caption{A ground spacecraft mounted on a manipulator.}
\label{fig_GroundSat}
\end{figure}

In this paper, we propose a method to control a manipulator with a
heavy payload, e.g., a test spacecraft, so that the closed-loop
system dynamics with respect to external force be as if the payload
is with inertia properties corresponding to a flight spacecraft \cite{Aghili-Namvar-2008}.
Fig. \ref{fig_GroundSat} schematically illustrates the test
spacecraft, i.e., a scale model of the flight spacecraft, is rigidly
attached to a manipulator arm. A six-axis force-moment sensor is
installed at the interface of the spacecraft and the manipulator,
for sensing  the external forces -- for instance, firing thrusters
-- superimposed by gravitational and inertial forces. Upon
measurement of the wrist force-moment and the joint angles and
velocities, the signals are used by a control system that moves the
manipulator and the test spacecraft with it appropriately. Such a
setup allows virtually testing the actual control system,
electronics, sensors, and actuators of a spacecraft in a closed-loop
configuration in the laboratory environment. The distinct
contribution of this work is a control system which incorporates
dynamics models of the test spacecraft (payload), flight spacecraft
as well as the manipulator to accurately replicate the motion
dynamics of the flight spacecraft using a scaled mockup, as
presented in Section~\ref{sec:cntr_law}. Notably, the controller can
compensate for the inertial forces of the payload without needing
any acceleration measurement; this is not attainable with the
conventional admittance controllers. A calibration procedure to
precisely null out the static component of the F/M sensor in
addition to sensitivity analysis are presented in
Section~\ref{sec:calibration}. Section~\ref{sec:flexible} is devoted
to emulation of spacecraft having flexible appendages, e.g., solar
panels.

\section{Control System}
\label{sec:cntr_sys}

\subsection{Dynamics Model}

The translational and the rotational motion dynamics of a {\em
flight spacecraft} can be conveniently expressed in a body-fixed
frame $\{ C_s \}$ as
\begin{equation} \label{eq_FlightDynamics}
M_s \dot{ \nu}_s + h_s(\nu_s) = {\cal F}_{\rm ext},
\end{equation}
where
\[M_s = \mbox{diag}\{  m_s I , \; I_{C_s} \} \;, \;\;\;\;\
h_s(\nu_s)= \begin{bmatrix} m_s \omega_s \times v_s \\
\omega_s \times I_{C_s} \omega_s \end{bmatrix} ,\] $I$ denotes the
identity matrix, $m_s$ and $I_{C_s}$ are the spacecraft mass and
inertia tensor, $\nu_s^T=[v_s^T \; \omega_s^T]$ is the generalized
velocity including the components of the linear velocity $v_s$ and
angular velocity $\omega_s$ of the spacecraft CM, and ${\cal F}_{\rm
ext}$ is the generalized external forces (due to the spacecraft
actuators, e.g., thrusters or reaction wheels). It is worth
mentioning that estimation of the other sources of external forces
and torques such as gravity gradient, thin-air drag, and solar
pressure can be added to the right-hand side (RHS) of
\eqref{eq_FlightDynamics} to achieve a more accurate result.

Fig. \ref{fig_GroundSat} illustrates the {\em test spacecraft} held
by a manipulator. The test spacecraft is of mass and inertia $m_m$
and $I_{C_m}$, respectively, that are different from those of the
flight spacecraft. The F/M sensor installed in the mechanical
interface of the manipulator and the test spacecraft allows us to
measure the force/moment interactions between the two systems. The
coordinate-frame $\{W \}$ is fixed to the manipulator base and, the
origin of body-fixed frame $\{C\}$ is chosen to be coincident with
the CM of the test spacecraft, and its orientation with respect to
frame $\{W\}$ is represented by the rotation matrix $ R$. The test
spacecraft is exposed to three different forces: the external force
${\cal F}_{\rm ext}$, gravitational force ${\cal F}_g$, and force
interaction between the test spacecraft and the manipulator ${\cal
F}_s$ that is measured by the F/M sensor. Note that ${\cal F}_s$ is
expressed in the body-fixed coordinate frame $\{ S \}$ coincident
with the sensor coordinate and parallel to $\{C\}$. Thus
\begin{equation} \label{eq_gravity}
{\cal F}_g = \begin{bmatrix} m_m g R^T k  \\
0
\end{bmatrix},
\end{equation}
where unit vector $k$ is aligned with the gravity vector
\footnote{If the z-axis of the coordinate frame $\{W\}$ is perfectly
parallel to the earth's gravity vector, then $k^T =[
\begin{array}{ccc} 0 & 0 & -1
\end{array} ]$.} which is expressed in the manipulator's base frame $\{ W \}$, and
$g=9.81 \; \mbox{m/s}^2$.  Similar to \eqref{eq_FlightDynamics}, the
dynamics of the test spacecraft can be described by
\begin{align} \notag
M_m \dot{ \nu} + h_m({ \nu}) & = -T {\cal F}_s + {\cal F}_g + {\cal
F}_{\rm ext} \\ \label{eq_MockpDynamics} &= -{\cal F}_{sg} + {\cal
F}_{ext} .
\end{align}
where $T$ denotes the transformation from frame $\{S\}$ to $\{C\}$,
i.e.
\[T = \begin{bmatrix} I & 0 \\ - [{ c} \times] & I \end{bmatrix}, \quad  [{ c}\times]=
\begin{bmatrix} 0 & -c_z & c_y \\ c_z & 0 & -c_x \\ -c_y & c_x &
0 \end{bmatrix}, \] vector $c$ denotes the location of the
center-of-mass and
\begin{equation} \label{eq_f'_s}
{\cal  F}_{sg} \triangleq T {\cal F}_s - {\cal F}_g .
\end{equation}

\subsection {Control Law} \label{sec:cntr_law}

We assume that both the test and flight spacecraft experience the
same actuation force ${\cal F}_{\rm ext}$, and that their
generalized velocities are the same, i.e., $\nu=\nu_s$. Under these
assumptions, we can say the test spacecraft is dynamically
equivalent to the flight spacecraft if they produce identical
accelerations, i.e., $\dot \nu = \dot \nu_s$. However, the
accelerations are governed by two different equations of motion, and
hence, in general, $\dot{ \nu} \neq \dot \nu_s$. Nevertheless, it is
possible to achieve dynamical similarity if the manipulator is
properly controlled. To this end, we define an estimation of the
acceleration $\dot{ \nu}^{\star}$ that is obtained by subtracting
\eqref{eq_MockpDynamics} from \eqref{eq_FlightDynamics}, i.e.,
\begin{subequations}
\begin{equation} \label{eq_DeltaDynamics}
M_{\Delta} \dot \nu ^{\star} + h_{\Delta}  = {\cal F}_{sg},
\end{equation}
where
\begin{align} \label{eq_M_Delta}
M_{\Delta} & \triangleq \begin{bmatrix} (m_s- m_m) I &  0 \\
0 & I_{C_s} -  I_{C_m} \end{bmatrix}, \\ \label{eq_h_Delta}
h_{\Delta} & \triangleq  \begin{bmatrix} (m_s-m_m) \omega \times v \\
\omega \times (I_{C_s} - I_{C_m}) { \omega}
\end{bmatrix}.
\end{align}
\end{subequations}

\begin{assumption} \label{invertible_M}
In the followings, we assume that  $M_{\Delta}$ is a non-singular
matrix, i.e.,
\begin{equation} \label{eq:nonequal_inertia}
m_s \neq m_m \quad  \text{and} \quad \lambda_i(I_{C_s}-I_{C_m}) \neq
0 \quad \forall i=1, \cdots 3.
\end{equation}
\end{assumption}

Notice that $\dot \nu^{\star}$ does not have any physical meaning,
rather it is just a definition. Let $J=\begin{bmatrix} J_v^T &
J_{\omega}^T \end{bmatrix}^T$ represent the manipulator Jacobian
expressed in the coordinate frame $\{C\}$, where sub-matrices $J_v$
and $J_{\omega}$ denote the translational and rotational Jacobians,
respectively. That is $v(q,\dot q) = J_v \dot q$ and $\omega(q,\dot
q ) = J_{\omega} \dot q$, where $q$ is the vector of joint angles.
The time derivative of the velocity equation leads to
\begin{equation} \label{eq_ddq}
\dot \nu = J \ddot q + \dot J \dot q.
\end{equation}
In view of equations \eqref{eq_ddq} and \eqref{eq_DeltaDynamics} and
Assumption~\ref{invertible_M}, we define $\ddot q^{\star} $ to be an
estimation of the joint accelerations as
\begin{subequations}
\begin{align} \label{eq_hatddq1}
\ddot{ q}^{\star} & \triangleq J^{-1}(\dot{ \nu}^{\star} - \dot{ J}
\dot{ q}) \\  \label{eq_hatddq} & = J^{-1} { M}_{\Delta}^{-1} {\cal
F}_{sg} - J^{-1} \big( N + \dot J \big) \dot q ,
\end{align}
\end{subequations}
with $M_{\Delta}^{-1} h_{\Delta} = N \dot q$ and
\[ N(q , \dot q) \triangleq  \begin{bmatrix} [J_{\omega} \dot{ q} \times] J_v
\\ (I_{C_s} - I_{C_m})^{-1} [J_{\omega} \dot{ q} \times ](I_{C_s} - I_{C_m})
J_{\omega}  \end{bmatrix}. \] Note that \eqref{eq_hatddq} is
obtained assuming that kinematic singularity does not occur.

Assume that the manipulator dynamics are characterized by inertia
matrix $M_r(q)$ and the nonlinear vector $h_r(q, \dot q)$, which
contains Coriolis, centrifugal and gravitational terms. One can show
that the equations of motion of the combined system of the
manipulator and the payload can be written in the standard form as:
\begin{subequations}
\begin{equation} \label{eq_DynRobot}
M_t \ddot{ q} + h_t(q, \dot q) = \tau + J^T {\cal F}_{\rm ext}
\end{equation}
where $\tau$ denotes the joint torques, and
\begin{align}
M_t(q) & \triangleq J^T M_m J + M_r(q), \\
h_t(q, \dot q) & \triangleq h_r(q, \dot q) + J^T h_m(\dot q) + J^T
M_m \dot J \dot q - m_m g J_v^T R^T k
\end{align}
\end{subequations}
Now, the
objective is to force the manipulator to follow the trajectory
dictated by \eqref{eq_hatddq}. Although it seems that this goal can
be achieved by using an inverse-dynamics controller
\cite{Spong-Vidasagar-1989,Canudas-Siciliano-Bastin-book-1996,Aghili-2010} based
on the manipulator dynamics, such a controller
will lead to an algebraic loop that is not legitimate from the
control point of view. Note that the force sensor signal contains
components of the inertial forces due to the acceleration. Thus,
compensating for ${\cal F}_s$ results in a torque control law which
has a direct component of the acceleration, while the acceleration
is algebraically related to the joint torques. This problem can be
alleviated by using an inverse-dynamics controller based on the
complete model \eqref{eq_DynRobot} and by compensating for an
estimation of the external force ${\cal F}_{\rm ext}$. That is
\begin{align} \notag
\tau & = M_t(q) \ddot q^{\star} + h_t(q, \dot q) - J^T {\cal F}^{\star}_{\rm ext} \\
\label{eq:tau1} & + M_t(q) \Big( K_d \big(\int \ddot q^{\star} dt -
\dot q \big) + K_p \big( \int \int \ddot q^{\star} dt - q \big)
\Big),
\end{align}
with $K_d= k_d I$ and $K_p = k_p I$ being the controller gains and
${\cal F}^{\star}$ being an estimation of the external force. In the
following analysis, we will show that the above  inverse-dynamics
controller in conjunction with a force estimator lead to exponential
stability. Let $\ddot {\tilde q}\triangleq \ddot q^{\star} - \ddot
q$ denotes the joint acceleration error, then the corresponding
Cartesian acceleration error is readily obtained from definition
\eqref{eq_hatddq1} as
\begin{equation} \label{eq:tilde_nu}
\dot {\tilde \nu } \triangleq  \dot \nu^{\star} -\dot \nu = J(q)
\ddot {\tilde q}.
\end{equation}
Substitution of $\dot \nu ^{\star}$ obtained from
\eqref{eq_DeltaDynamics} into the above equation yields
\[ \dot \nu = M_{\Delta}^{-1} {\cal F}_{sg} - N \dot q - J \ddot{\tilde q}. \]
Now, upon substitution of the acceleration from the above into
\eqref{eq_MockpDynamics}, we can write the expression of the
external force as:
\[ {\cal F}_{ext} =  {\cal F}_{\rm ext}^{\star} + \tilde {\cal F}_{\rm ext},\]
where
\begin{equation} \label{eq:F_est}
{\cal F}_{\rm ext}^{\star} = \big( I + M_m M_{\Delta}^{-1} \big)
{\cal F}_{sg} + h_m - M_m N \dot q
\end{equation}
is the estimation of the external force and
\begin{equation} \label{eq:tilde_F}
\tilde {\cal F}_{ext} = - M_m J \ddot {\tilde q}
\end{equation}
is the force estimation error. Clearly, the force estimation error
goes to zero only if the acceleration error does so. We will show
that under a mid condition, controller \eqref{eq:tau1} in
conjunction with force estimator \eqref{eq:F_est} results in
exponential stabling of the motion and force errors. To this end,
substitution of ${\cal F}^{\star}_{\rm ext}$ and $\ddot q^{\star}$
obtained from \eqref{eq:F_est} and \eqref{eq_hatddq}, respectively,
into \eqref{eq:tau1} yields the expression of the control law as:
\begin{align} \notag
\tau & = J^T \big( M_{Cr}(q) M_{\Delta}^{-1} - I \big) {\cal F}_{sg}
+ h_r(q, \dot q)  - M_r(q) J^{-1}(N(q,\dot
q) + \dot J) \dot q  \\
\label{eq_Control} & - m_m g J_v^T R^T k + M_t(q) \Big( K_d \big(
\int \ddot q^{\star} \; \mbox{d} t - \dot q \big) + K_p \big( \int
\int \ddot q^{\star} \; \mbox{d}t - q \big) \Big),
\end{align}
where $ M_{Cr} \triangleq J^{-T} M_r J^{-1}$  is the {\em Cartesian
inertia} of the manipulator. Stability of closed-loop system remains
to be proved. Knowing that \eqref{eq_Control} becomes equivalent to
\eqref{eq:F_est} if the force term, ${\cal F}_{\rm ext}$, of the
former equation is replaced by ${\cal F}^{\star}_{\rm ext}={\cal
F}_{\rm ext} - \tilde {\cal F}_{\rm ext}$, we can arrive at the
equations of the motion and force errors by substituting
\eqref{eq_Control} into system \eqref{eq_DynRobot}, i.e.,
\begin{equation*}
M_t \big( \ddot {\tilde{q}} + K_d \dot{\tilde{q}} + K_p {\tilde{q}}
\big) = - J^T \tilde {\cal F}_{\rm ext}.
\end{equation*}
Moreover, we know that the force and acceleration errors are related
by \eqref{eq:tilde_F}. Thus
\begin{equation*}
M_r \ddot {\tilde{q}} + M_t \big( K_d \dot{\tilde{q}} + K_p
{\tilde{q}} \big)= 0,
\end{equation*}
which can be rewritten as:
\begin{equation} \label{eq_accODE}
\ddot {\tilde{q}} +  K_d \dot{\tilde{q}} + K_p {\tilde{q}} + Q(q)
\big( K_d \dot{\tilde{q}} + K_p {\tilde{q}}  \big)= 0,
\end{equation}
where
\begin{equation} \label{eq:Q}
Q \triangleq M_r^{-1} \big(J M_m J^T \big).
\end{equation}

We will show that system \eqref{eq_accODE} remains stable if the
coefficient matrix of the additive term, $Q$, is sufficiently small.
Let assume that $x^T=[\tilde q^T \quad \dot{\tilde q}^T]$ represent
the sate vector. Then, \eqref{eq_accODE} can be written as
\begin{equation} \label{eq:perturbed}
\dot x = A x + \epsilon(t, x)
\end{equation}
where
\[ A=\begin{bmatrix} 0 & I \\ -K_p & -K_d \end{bmatrix} \quad \text{and}
\quad \epsilon(t, x)=- Q \begin{bmatrix} 0 \\ K_p \tilde q + K_d
\dot{\tilde q} \end{bmatrix}.\] Since the perturbation term
$\epsilon$ satisfies the linear growth bound
\[ \| \epsilon \| \leq  \sqrt{k_p^2 + k_d^2} \|Q \| \| x \|, \]
system \eqref{eq:perturbed} is in the form of {\em vanishing
perturbation} \cite{Khalil-1992}. Moreover, since $A$ is Hurwitz,
there exists Lyapunov function
\begin{equation} \label{eq:Lyapunov}
V(x) = x^T P x
\end{equation}
with $P> 0 $ satisfying
\begin{equation} \label{eq:PA}
PA + A^T P = -I.
\end{equation}
The derivative of $V(x)$ along trajectories of perturbed system
\eqref{eq:perturbed} satisfies
\begin{equation}
\dot V \leq \big(-1  + 2\sqrt{k_p^2 + k_d^2} \lambda_{\rm max}(P)
\|Q \| \big) \| x \|^2
\end{equation}
On the other hand, the solution of the Lyapunov equation
\eqref{eq:PA} is given by
\[ P = \frac{1}{2k_pk_d} \begin{bmatrix} k_p(k_p +1 ) + k_d^2 & k_d
\\ k_d & k_p+1 \end{bmatrix}, \]
which verifies
\[ \lambda_{\rm max}(P) \leq  \frac{(k_p+1)^2 + k_d^2}{2k_p k_d}. \]
Therefore, according to the stability theorem of perturbed system
\cite[p. 206]{Khalil-1992}, the origin of \eqref{eq:perturbed} is
globally exponentially stable if
\begin{equation} \label{eq:alpha}
\| Q \| \leq \alpha(k_p,k_d) = \frac{k_p k_d}{\big((k_p+1)^2 + k_d^2
\big)^{\frac{3}{2}}}.
\end{equation}
Using the norm properties in \eqref{eq:Q}, we obtain a conservative
condition for the stability as:
\begin{equation} \label{eq:mass_inequality}
\lambda_{\rm max}(M_m) \leq  \alpha(k_p,k_d) \frac{\lambda_{\rm
min}(M_r)}{\lambda_{\rm max}(JJ^T)}.
\end{equation}

Now, if \eqref{eq:alpha} is satisfied, then there must exist scalar
$\Omega >0$ such that $\| x \| \leq \| x(0) \| e^{-\Omega t}$.
Therefore, it can be inferred from \eqref{eq_accODE} that
\begin{equation} \label{eq_bound_ddq}
\| \ddot{\tilde{q}} \| \leq a e^{-\Omega t},
\end{equation}
where $a =(k_p^2 + k_d^2)(1+ \| Q \|) \| x(0) \|$.

Now, we are ready to derive the input/output relation of the closed
loop system under the proposed control law. Adding both sides of
\eqref{eq_MockpDynamics} and \eqref{eq_DeltaDynamics} yields
\begin{equation} \label{eq:ddotv-ddotv*}
M_s \dot \nu + M_{\Delta} \dot \nu^{\star} + h_s = {\cal F}_{\rm
ext}.
\end{equation}
Finally, using \eqref{eq:tilde_nu} in \eqref{eq:ddotv-ddotv*}, the
equations of motion of the test spacecraft become
\begin{subequations}
\begin{equation} \label{eq_purturbed}
M_s \dot \nu + h_s(\nu) = {\cal F}_{\rm ext} + \delta,
\end{equation}
where
\begin{equation} \label{eq_delta}
\delta(t)= M_{\Delta} J \ddot{\tilde q}
\end{equation}
\end{subequations}
is a non-vanishing perturbation. Since $J$ is always a bounded
matrix, we can say
\[ \sigma = \max_q \sqrt{\lambda_{\rm max}({ J}^T { J})}, \] where
$\lambda_{\rm max}(\cdot)$ denotes the maximum eigenvalue of a
matrix. It follows from \eqref{eq_bound_ddq}  and \eqref{eq_delta}
that
\begin{equation} \label{eq_del}
\| \delta \| \leq  \sigma a \lambda_{\rm max}({ M}_{\Delta})
e^{-\Omega t},
\end{equation}
which means that the perturbation exponentially  relaxes to zero
from its initial value. The above development can be summarized in
the following.

\begin{proposition}
Let a rigid-body object with generalized inertia $M_m$ attached to a
manipulator with inertia $M_r$. Assume that the force/moment
developed at the interface of the object and the manipulator is
sensed and fed back to the manipulator according to the control law
\eqref{eq_Control}. Moreover, assume that
\eqref{eq:nonequal_inertia} and \eqref{eq:alpha} are satisfied.
Then, the motion of the object in response to external force ${\cal
F}_{\rm ext}$ obeys equation of motion of another rigid-body object
characterized by generalized inertia $M_s$.
\end{proposition}

\subsection{Force Feedback Gain} \label{sec:boundedness}

Ideally, the controller of the emulating system can change the
inertia of the test spacecraft to any desired value. However, there
are constraints  \eqref{eq:nonequal_inertia} and
\eqref{eq:mass_inequality} on the inertia matrices of the test and
flight spacecraft as well as the manipulator that must be considered
in the design. Assuming a steady-state mode in which the control
error reaches zero, we can express the torque-control input by
\begin{equation} \label{eq_alpha}
\tau = J^T (M_{Cr} { M}_{\Delta}^{-1} - I) {\cal F}_{sg} + \eta (q ,
\dot{ q}),
\end{equation}
where $ \eta(q, \dot q)$ represents the motion dependent portion of
the feedback, while the first term in the RHS of equation
\eqref{eq_alpha} is force feedback. In the following we examine two
extreme cases of the force feedback gain.

\subsubsection{Zero Gain}
Equation (\ref{eq_alpha}) implies that the force feedback is
disabled if $M_{Cr}=M_{\Delta}$ or
\begin{equation} \label{eq:equal_mass}
M_{Cr}(q)+ M_m=M_s.
\end{equation}
Clearly, to implement the emulation controller without force
feedback requires satisfying \eqref{eq:equal_mass} for all possible
postures. However, with the exception of Cartesian manipulators,
most manipulators are of configuration-dependent inertia matrix,
whereas the spacecraft inertia are constant matrices. This means
that the condition \eqref{eq:equal_mass} can be satisfied only for
few isolated postures at best. It is worth mentioning that the case
of $M_s=M_m$ becomes a favorable condition if $M_{C_r}\equiv 0$,
i.e., the manipulator inertia is negligible; see
\eqref{eq:equal_mass}. However, a manipulator with zero mass (and
zero joint friction) can be though of as an air-bearing simulator
system, which has its own shortcomings as descried in
Section~\ref{sec_introduction}.

\subsubsection{Infinite Gain}

It is apparent from (\ref{eq_alpha}) that for the control torque
effort to be bounded requires that $M_{\Delta}$ be a non-singular
matrix, i.e., condition \eqref{eq:nonequal_inertia} is satisfied. At
first glance, this result seems counterintuitive. But, it can be
seen from \eqref{eq_DeltaDynamics} that the acceleration and thus
the subsequent motion trajectory can be uniquely estimated only if
$M_{\Delta}$ is a full-rank matrix. It is also apparent from
\eqref{eq_FlightDynamics} and \eqref{eq_MockpDynamics} that the only
possibility for the  flight and test spacecraft with the same mass
and inertia to produce similar velocity and acceleration
trajectories with respect to external force ${\cal F}_{ext}$ is that
the interaction force ${\cal F}_{sg}$ becomes zero. Clearly, in such
as case, it is not possible to predict the position and velocity
trajectories from the estimated acceleration and hence the feedback
is meaningless.

\section{Simulating a Micro-G Environment}
\label{sec:calibration}

\subsection{Precise Gravity Compensation}

Performing a high-fidelity zero-g emulation critically relies on a
precise force/moment feedback which, in turn, is determined by: (i)
Accuracy of the gravity compensation; (ii) the resolution of the F/M
sensor. These issues are discussed below.

\subsubsection{Calibration}

The static components of the F/M sensor output include the sensor
offset and the payload gravitational force, which are not
distinguishable from each other. Nevertheless, if a sequence of
sensor readings is recorded by locating the manipulator in several
known poses, it is possible to identify the sensor offset together
with all the gravity parameters that are required to null out the
static components of the sensor.

If the gravity were completely compensated, then for every position
we would have ${\cal F}_{sg}=0$, i.e., $T({\cal  F}_s -{\cal F}_0)
-{\cal F}_g= 0$, where ${\cal F}_0^T=\begin{bmatrix} f_0^T & n_0^T
\end{bmatrix}$ denotes the sensor offset. Now, we
consider $\{{\cal F}_0,m_m, c, k\}$ as the set of uncertain
parameters that are be identified. Defining vector $w \triangleq m_m
k$ and knowing that $[c \times] R^T w = -[(R^T w) \times] c$, we can
break up ${\cal F}_s=T^{-1} {\cal F}_g + {\cal F}_0$ into two {\em
linear regression} equations as
\begin{subequations}
\begin{align} \label{eq_Fs}
f_s & =   \begin{bmatrix}  I & g{ R}^T
\end{bmatrix} \begin{bmatrix} f_0 \\ w \end{bmatrix},\\ \label{eq_Ns}
n_s & =  \begin{bmatrix} I & -gm_m[(R^T k) \times]
\end{bmatrix}  \begin{bmatrix} n_0 \\ c \end{bmatrix},
\end{align}
\end{subequations}
where ${\cal F}_s^T=\begin{bmatrix} f_s^T & n_s^T
\end{bmatrix}$. Now stacking $p$ measurements $y_1^T=[f_{s1}^T,
f_{s2}^T, \cdots, f_{sp}^T ]$ and $y_2^T=[n_{s1}^T, n_{s2}^T,
\cdots, n_{sp}^T ]$, that are obtained by configuring the
manipulator at $p$ different positions $\{ q_1, q_2, \cdots, q_p\}$,
we can derive two linear matrix relation $y_1 = \Psi_1(q) {
\Theta}_1$ and $y_2 = \Psi_2(q,\Theta_1) \Theta_2$ from
(\ref{eq_Fs}-\ref{eq_Ns}), where vectors $\Theta_1$ and $\Theta_2$
contain the parameters of interest. Finally, assuming a sufficient
number of independent equations, one can obtain the vectors of
estimated parameters $\hat{ \Theta}_1$ and $\hat \Theta_1$
consecutively by using the least squares method from
\[ \hat \Theta_1 = \Psi_1^+ y_1, \;\;\; \mbox{and} \;\;\;
\hat \Theta_2 = \Psi_2^+(\hat \Theta_1) y_2,\] where $\Psi_i^+={
\Psi}_i^T ( \Psi_i \Psi_i^T)^{-1}$ is the pseudo-inverse of
$\Psi_i$. Note that the mass and the gravitational vector can be
retrieved from
\[\hat m_m= \| \hat w \| \quad \text{and} \quad  \hat k  = \frac{\hat w}{ \| \hat w \|} .\]

\subsubsection{Position Errors and Accuracy of the Gravity Compensation}
Error between the measured joint angles used by the gravity
compensator and the true joint angles will result in a small
residual static force acting on the payload. One source of this
error is measured quantization. In order to minimize the residual
force induced by the quantization as much as possible, we need to
employ high-resolution encoders at the joints so that the induced
error becomes at least comparable to the F/M sensor resolution. In
the following we relate the errors in the gravity compensation and
the that of joint angles.

Assume that $\Delta q$ and  $\Delta f_s$ denote small errors in
measured joint angles and the computed gravity force, respectively.
Using the Taylor series of \eqref{eq_Fs} leads to
\begin{equation} \label{eq:taylor}
\| \Delta f_s \| \leq g m_m  \left \| \frac{\partial}{\partial q}
R^T(q) k \right \| \| \Delta q \|.
\end{equation}
Using the facts that all elements of the rotation matrix are
sinusoidal functions of $q$ and that $k$ is a unit vector, one can
show that a conservative bound on the first norm of the RHS of
\eqref{eq:taylor} is 6. Thus, a bound on the force error can be
found as
\begin{equation} \label{eq:delta_f}
\| \Delta f_s \| \leq 6 g m_m  \| \Delta q. \|
\end{equation}
Similar argument shows that a conservative bound on the magnitude of
moment error $\Delta n_s$ can be found as
\begin{equation} \label{eq:delta_n}
\| \Delta n_s \| \leq 6 g m_m \| c \|  \| \Delta q. \|
\end{equation}

\subsection{Assessing the Quality of the Micro-Gravity Environment}
\label{sec:mug} Emulation in a zero-gravity environment requires the
static component of the F/M sensor is perfectly nulled out. However,
in practice, this requirement can not be completely satisfied due to
errors. A natural question rises; what is the quality of the
emulator in simulating a weightlessness environment? To answer this
question, let us assume that $\delta \bar f_{sg}$ denote the average
magnitude error of the compensated F/M sensor output over several
payload static poses. Then, the average acceleration introduced to
the emulating system can be simply obtained by dividing the
magnitude of this force by the inertia of the spacecraft being
simulated. Normalizing the acceleration w.r.t. the Earth gravity
constant, we define the following dimensionless index
\begin{equation} \label{eq:microg_index}
\gamma \triangleq  \frac{\bar{\delta f_{sg}}}{g m_s} \times 10^6 =
\frac{\| \sum_i^n f_{s_i} - \Psi_{1_i} \hat \Theta_1 \|}{n g m_s}
\times 10^6
\end{equation}
to measure the virtual gravity of the simulated environment. In
other words, the payload (test spacecraft) experiences as though it
moves under a  gravitational field with intensity of $\gamma \cdot g
$ rather than a zero-g environment. It is worth pointing out that
$\gamma$ can be also interpreted as the drift exhibited by the
emulation system. Similarly, the micro-gravity environment for the
rotational motion can be defined as
\[ \frac{\| \sum_i^n n_{s_i} - \Psi_{2_i} \hat \Theta_2 \|}{n g \| c \| m_m}
 \times 10^{6}.
\]

\begin{figure}
\centering
\includegraphics[width=8cm]{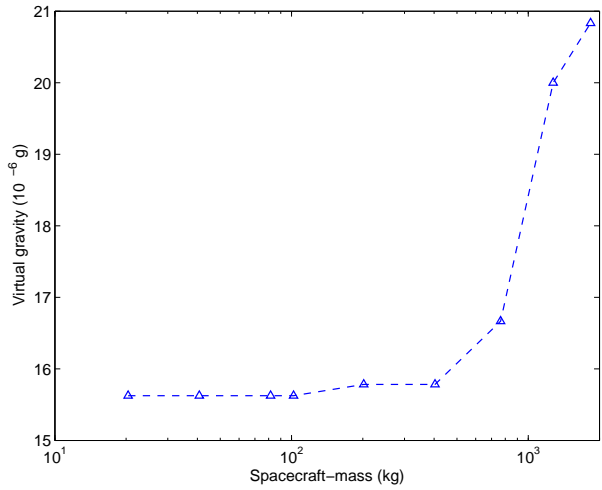}
\caption{The best achievable micro-g environment, computed from
the resolutions of a series of commercial F/M sensors, for emulation of spacecraft
with different masses.}\label{fig:microg}
\end{figure}

\subsubsection{Resolution of F/M Sensor}
At best, the force error $\delta \bar f_{sg}$ can be reduced down to
the resolution of the F/M sensor. The resolution of a F/M sensor
depends on its sensing range; a sensor with large sensing range
tends to have lower resolution and vice versa. Since the F/M sensor
is located at the manipulator-payload interface, the sensor should
be selected so that its sensing range matches the weight of payload,
i.e., the test spacecraft. Therefore, the ratio of the sensor
resolution to its sensing range is the emulation system limitation
in achieving the lowest micro-g.

Fig.~\ref{fig:microg} illustrates the best achievable micro-g's
versus different spacecraft masses that is calculated from the
resolutions and the sensing ranges of the commercial ATI F/M sensors
\cite{ATI-F/T-2007}. Here, we assume that the scaling factor of the
emulated spacecraft is two. It is evident from the figure that in
the emulation of small to medium size spacecraft with mass of up to
$500$kg, the sensor resolution is sufficient for achieving accuracy
of $16 \times 10^{-6}g$ (it almost remains constant in that range).
However, the value of the virtual gravity dramatically increases,
when the spacecraft mass exceeds that critical mass. This is due to
the fact that commercial F/M sensors with large load capacity come
with relatively low resolution. In order to improve the $\gamma$
factor, one may use a mechanism to counter the effects of gravity in
rigid-bodies
\cite{White-Yangsheng-1994,Rahman-Ramanathan-Seliktar-Harwin-1995,Gopalswamy-Gupta-Vidyasagar-1996,Ulrich-Kumar-1991}.
For example, using a passive counterweight
\cite{White-Yangsheng-1994} can substantially reduce the static load
on the F/M sensor, thereby allowing smaller and more precise sensor
to be selected. The main disadvantage of this method is introduction
of additional inertia. However, this is not an issue here because
the controller can scale  the inertia of the payload down or up to
any desired value.
\begin{figure}[t]
\centering \includegraphics[width=10cm]{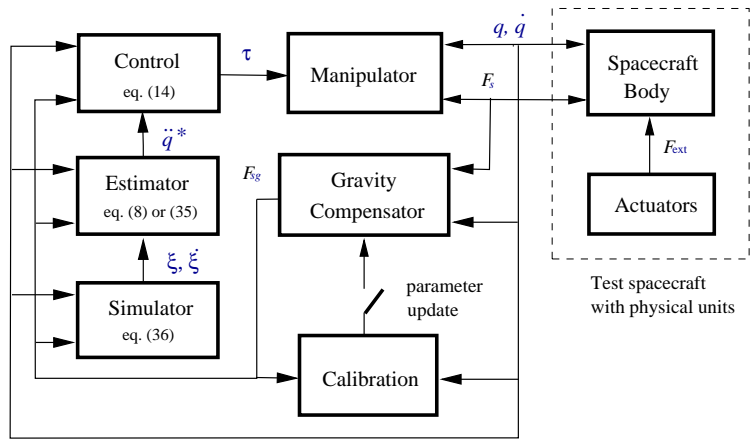}
\caption{Architecture of the ground testbed for emulation of
spacecraft.} \label{fig_Architecture}
\end{figure}
\section{Emulation of Flexible Spacecraft}
\label{sec:flexible}

Many spacecraft have flexible appendages, e.g. satellites with solar
panels, that can significantly affect their dynamics. However,
testing a flexible spacecraft in a 1-g environment poses many
difficulties due to large deformation induced by gravity. Indeed,
the structure of a solar panel cannot even hold itself against
gravity when it is fully deployed. Moreover, the location of the CM
of a flexible spacecraft is no longer fixed as it depends on the
flexural coordinates whose direct measurement is not usually
available. In the following, we extend the emulation concept for the
case where the test spacecraft is  rigid while the target flight
spacecraft is flexible.  It is assumed that the actuators are
mounted to the rigid part of the test spacecraft. The test
spacecraft lacks any flexible hardware, such as solar panels. Yet,
motion perturbation caused by the flexible appendages is generated
by simulation and then superimposed on the trajectories that
subsequently drive the manipulator.

Let $\xi$  denote the flexural coordinates of a flexible spacecraft.
Then, the equations of motion for the entire system can be written
in the partitioned mass matrix form
\begin{equation} \label{eq_FlexDynamics}
\begin{bmatrix} M_s & M_{sf} \\ M_{sf}^T & M_f
\end{bmatrix}  \begin{bmatrix} \dot \nu \\ \ddot
\xi \end{bmatrix}  +  \begin{bmatrix} h_{sr}({\nu}, \xi, \dot \xi)
\\  h_{sf}({\nu}, \xi, \dot \xi)
\end{bmatrix}  =  \begin{bmatrix} {\cal F}_{\rm ext} \\ 0 \end{bmatrix},
\end{equation}
where $M_f$ is the flexural inertia matrix, $M_{sf}$ is the cross
inertia matrix, $h_{sr}$ and $h_{sf}$ are the nonlinear vectors
associated with the rigid and flexural coordinates. Analogous to the
case of rigid spacecraft, subtracting equation
\eqref{eq_FlexDynamics} from \eqref{eq_MockpDynamics} eliminates
${\cal F}_{\rm ext}$ from the the equations of motion. Defining
$\bar M_{\Delta}= M_{\Delta} - M_{sf} M_f^{-1} M_{sf}^T$ and
$h_{\Delta}= h_{sr} - h_m$, we can write the accelerations of the
rigid and the flexural coordinates  by
\begin{equation} \label{eq_RigidInertia}
\ddot q^{\star} = - J^{-1} ( N + \dot J) \dot q - J^{-1} \bar
M_{\Delta}^{-1} M_{sf} M_f^{-1} h_{sf} + \bar M_{\Delta}^{-1}{\cal
F}_{sg},
\end{equation}
and
\begin{equation} \label{eq_ddxi}
\ddot \xi = - M_f^{-1}(I + M_f^{-1} M_{sf}^T M_{sf} M_f^{-1}) h_{sf}
- M_f^{-1} M_{sf}^T \bar M_{\Delta}^{-1} ({\cal F}_{sg} -
h_{\Delta}).
\end{equation}
Equation \eqref{eq_RigidInertia} can be substituted in
\eqref{eq_Control} to obtain the control law. However, to calculate
the acceleration from \eqref{eq_RigidInertia} requires the value of
the flexural states because $h_{sr}$  and $h_{sf}$ are functions of
$\xi$ and $\dot \xi$. An estimation of the flexural states can be
obtained by simulation. First, the acceleration of the flexural
coordinate can be computed by making use of the acceleration model
(\ref{eq_ddxi}), and then the flexural states are obtained as a
result of numerical integration.

The architecture of the zero-g emulating testbed for spacecraft is
illustrated in Fig.~\ref{fig_Architecture}. To summarize, the
emulation of flexible spacecraft may proceed as the following steps:
\begin{enumerate}
\item start at a time when all of the system states, i.e., $\{ q,
\dot q, \xi, \dot \xi \}$ are known,
\item  estimate the joint acceleration from (\ref{eq_RigidInertia}) (use \eqref{eq_hatddq}
instead for rigid spacecraft), \label{step_estimation}
\item apply control law \eqref{eq_Control} to the manipulator;
\item  obtain the flexural
states as a result of the consecutive integration of the
acceleration obtained from \eqref{eq_ddxi} -- skip this step for
rigid spacecraft -- and then go to step \ref{step_estimation}.
\end{enumerate}

\begin{figure}[t]
\centering
\includegraphics[width=11cm]{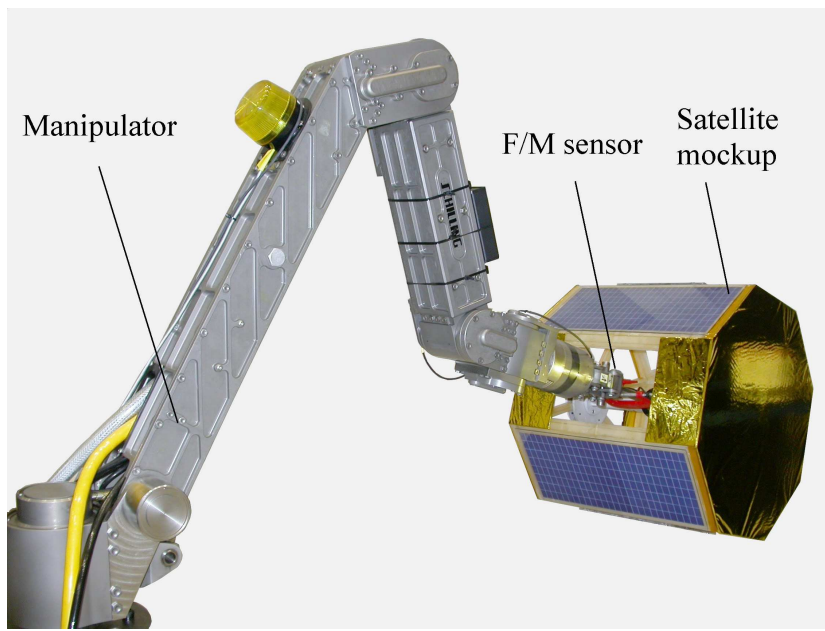}
\caption{The spacecraft simulator using actively controlled manipulator.}
\label{fig_Schilling}
\end{figure}

\section{Conclusions}

A control system for a manipulator carrying a rigid-body payload has
been developed in order to modify the motion dynamics of the
combined system in consequence of external according to that of a
free-floating body which has different inertial properties from the
payload. This allowed zero-g emulation of the scaled prototype of a
spacecraft (with non-negligible inertia) in a 1-g laboratory
environment. It was shown that the controller in conjunction with
the motion and force estimators could drive the manipulator so as to
achieve dynamical similarity between the test and flight spacecraft.
Notably, the controller can compensate for the inertial forces of
the heavy payload (test spacecraft) without needing any acceleration
measurement.

The stability of the closed loop system was analytically
investigated. The results showed that system remains stable provided
that mass and inertia of the test and flight spacecraft are not the
same and that the norm of the inertia ratio of the payload to
manipulator is upper bounded by a scaler which is a function of the
controller gains. Finally, the methodology was extended for
emulation of spacecraft having flexible appendages, e.g. solar
panels.

A calibration procedure to precisely null out the static component
of the F/M sensor was developed that tunes the gravity, kinematic,
and sensor parameters all together. A sensitivity analysis showed
that the position and force sensors have to be with specified
resolutions in order to achieve a certain level of micro-gravity.

\bibliographystyle{IEEEtran}

\end{document}